\documentclass[10pt,twocolumn,letterpaper]{article}
\usepackage[accsupp]{axessibility}

\usepackage{iccv}
\usepackage{times}
\usepackage{epsfig}
\usepackage{graphicx}
\usepackage{amsmath}
\usepackage{amssymb}

\usepackage{url}            % simple URL typesetting
\usepackage{booktabs}       % professional-quality tables
\usepackage{amsfonts}       % blackboard math symbols
\usepackage{nicefrac}       % compact symbols for 1/2, etc.
\usepackage{microtype}      % microtypography
\usepackage{xcolor}         % colors
\usepackage{graphicx}
\usepackage{amsmath}
\usepackage{amssymb}
\usepackage{amsfonts}
\usepackage{multirow}
\usepackage{color}
\usepackage{colortbl}
\usepackage{wrapfig}
\usepackage{balance}
\usepackage{paralist}
\usepackage{authblk}

\definecolor{mygray}{gray}{.96}
\definecolor{myblue}{RGB}{240,248,255}
\definecolor{myred}{RGB}{255,228,225}

\DeclareMathOperator{\EM}{E_{LAC}} 
\DeclareMathOperator{\EV}{E_{V}} 
\DeclareMathOperator{\EVS}{E_{Vs}} 
\DeclareMathOperator{\EVF}{E_{Vf}} 
\DeclareMathOperator{\D}{D_{LAC}}

\def\ie{\emph{i.e.}}
\def\eg{\emph{e.g.}}

\def\wrt{\emph{w.r.t.}}

\DeclareMathOperator{\Sim}{Sim}

\usepackage[pagebackref=true,breaklinks=true,colorlinks,bookmarks=false]{hyperref}
\iccvfinalcopy % *** Uncomment this line for the final submission

\def\iccvPaperID{****} % *** Enter the ICCV Paper ID here

\ificcvfinal\fi

\begin{document}

%%%%%%%%% TITLE
\title{ Fine-grained Composable Skeleton Motion Representation Learning for Temporal Action Segmentation}
\title{ Fine-grained Skeleton Representation Learning via Latent Motion Navigation and Composition}
\title{ CoM: Composable Motion Synthesis for Fine-grained Skeleton Action Representation Learning}
\title{LAC - Latent Action Composition for Skeleton-based Action Segmentation}
\author{
    %Authors
    % All authors must be in the same font size and format.
    Di Yang\textsuperscript{\rm 1} \hskip 1em
    Yaohui Wang\textsuperscript{\rm 1\thanks{Corresponding author.}} \hskip 1em
    Antitza Dantcheva\textsuperscript{\rm 1} \hskip 1em
    Quan Kong\textsuperscript{\rm 3} \hskip 1em
    Lorenzo Garattoni\textsuperscript{\rm 2}  
    
    Gianpiero Francesca\textsuperscript{\rm 2} \hskip 1em 
    François Brémond\textsuperscript{\rm 1} %\hskip 1em
    %Afiliations
    \\
    \textsuperscript{\rm 1}Inria, Université Côte d'Azur \hskip 1em 
    \textsuperscript{\rm 2}Toyota Motor Europe \hskip 1em 
    \textsuperscript{\rm 3}Woven by Toyota \\ % \hskip 1em 
    
{\tt\small \{di.yang, yaohui.wang, antitza.dantcheva, francois.bremond\}@inria.fr} \hskip 1em

{\tt\small \{lorenzo.garattoni, gianpiero.francesca\}@toyota-europe.com \hskip 1em  
quan.kong@woven-planet.global}

\vspace{-0.cm}    
}

\maketitle
% Remove page # from the first page of camera-ready.
\ificcvfinal\thispagestyle{empty}\fi

%%%%%%%%% ABSTRACT

\begin{abstract}

Skeleton-based action segmentation requires recognizing composable actions in untrimmed videos. Current approaches decouple this problem by first extracting local visual features from skeleton sequences and then processing them by a temporal model to classify frame-wise actions. However, their performances remain limited as the visual features cannot sufficiently express composable actions.
%in untrimmed videos. 
In this context, we propose Latent Action Composition (LAC)\footnote{Project website: \url{https://walker1126.github.io/LAC/}}, a novel self-supervised framework aiming at learning from synthesized composable motions for skeleton-based action segmentation. LAC is composed of a novel generation module towards synthesizing new sequences. Specifically, we design a linear latent space in the generator to represent primitive motion. New composed motions can be synthesized by simply performing arithmetic operations on latent representations of multiple input skeleton sequences. {LAC leverages such synthesized sequences, which have large diversity and complexity, for learning visual representations of skeletons in both sequence and frame spaces via contrastive learning.} The resulting visual encoder has a high expressive power and can be effectively transferred onto action segmentation tasks by end-to-end fine-tuning without the need for additional temporal models. We conduct a study focusing on transfer-learning and we show that representations learned from pre-trained LAC outperform the state-of-the-art by a large margin on TSU, Charades, PKU-MMD datasets.
\end{abstract}

\section{Introduction}
\begin{figure*}
\begin{center}

\includegraphics[width=1\linewidth]{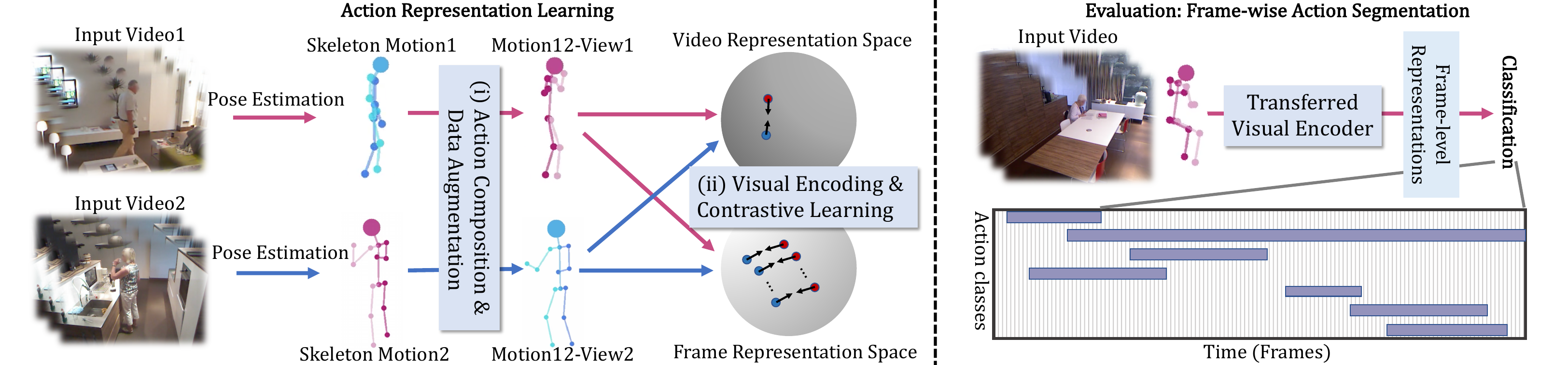}%0.96
\end{center}
   \vspace{-0.1cm}
   \caption{ \textbf{General pipeline of LAC.} Firstly, in the representation learning stage (left), we propose (i) a novel action generation module to combine skeletons of multiple videos (\eg, `Walking' and `Drinking' shown in the top and bottom respectively). 
   %and augmented by generations in different viewpoints. 
   We then adopt a (ii) contrastive module to pre-train a visual encoder by learning data augmentation invariant representations of the generated skeletons
   %(\ie, positive pairs) 
   in both video space and frame space. Secondly (right), the pre-trained visual encoder is evaluated by transferring to action segmentation tasks.}
\vspace{-0.1cm}%0.35
\label{fig:intro}
\end{figure*}

Human-centric activity recognition is a crucial task in real-world video understanding. In this context, \textit{skeleton data} that can be represented by 2D or 3D human keypoints plays an important role, as it is complementary to other modalities such as RGB~\cite{3dcnn, Carreira_2017_CVPR, 3d-resnet, slow-fast, x3d, Ryoo2020AssembleNetAM, li2021ctnet, Wang_2021_CVPR, arnab2021vivit}  and optical flow~\cite{two-stream, c-2stream}. As the human skeleton modality has witnessed {a tremendous boost in} robustness \wrt~content changes related to camera viewpoints and subject appearances, the study of recognizing activities directly from 2D/3D skeletons has gained increasing attention~\cite{Du2015HierarchicalRN, DING-jvcir, Caetano2019SkeletonIR, Yan2018SpatialTG, 2sagcn2019cvpr, aaai2021multiscale, res-gcn, unik, topology_2021_ICCV, graphscattering, duan2021revisiting, yang2022via}. While aforementioned approaches have achieved remarkable success, such approaches often focus on \textit{trimmed videos} containing \textit{single actions}, which constitutes a highly simplified scenario. Deviating from this, in this work, we tackle the challenging setting of \textit{action segmentation in untrimmed videos based on skeleton sequences}.

In untrimmed videos, activities are composable \ie, motion performed by a person generally comprises multiple actions (co-occurrence), each with the duration of a few seconds. Towards modeling \textit{long-term dependency} among different actions, expressive skeleton features are required. Current approaches~\cite{lea2017temporal, TGM, superevent, Dai_2022_PAMI} obtain such features through visual encoder such as AGCNs~\cite{2sagcn2019cvpr} pre-trained on trimmed datasets. 
{However, due to the limited motion information in the trimmed samples, the performance of such features in classifying complex actions is far from satisfactory. Towards addressing this issue, we propose to construct \textit{synthesized composable skeleton data} for training a more effective visual encoder, endowed with strong representability of subtle action details for action segmentation. 

In this paper, we propose Latent Action Composition (LAC), a novel framework aiming at leveraging synthesized composable motion data for self-supervised action representation learning. As illustrated in Fig.~\ref{fig:intro} (left), as opposed to current self-supervised approaches~\cite{lea2017temporal, TGM, superevent, Dai_2022_PAMI}, LAC learns action representations in two steps: a first \textit{action composition} step is then followed by a \textit{contrastive learning} step.

\textit{Action composition} is a novel initialization step to train a generative module that can generate new skeleton sequences by combining multiple videos. As high-level motions are difficult to combine directly by the joint coordinates (\eg, `drink' and `sitdown'), LAC incorporates a novel Linear Action Decomposition (LAD) mechanism within an autoencoder. LAD seeks to learn an action dictionary to express subtle motion distribution in a discrete manner. Such action dictionary incorporates an orthogonal basis in the latent encoding space, containing two sets of directions. The first set named {`Static'} includes directions representing static information of the skeleton sequence, \eg, viewpoints and body size. The other set named `Motion' includes directions representing temporal information of the skeleton sequence, \eg, the primitive dynamics of the action performed by the subject. The new skeleton sequence is generated via a linear combination of the learned {`Static'} and `Motion' directions. 
We adopt motion retargeting to train the autoencoder and the dictionary using skeleton sequences with `Static' and `Motion' information built from 3D synthetic data~\cite{mixamo}. 
Once the action dictionary is constructed, in the following \textit{contrastive learning} step, `Static'/`Motion' information and action labels are not required and composable motions can be generated from any multiple input skeleton sequences 
%with their respective `Static' parts 
by combining their latent `Motion' sets.% in the latent codes.   
%\item 

The \textit{contrastive learning} step aims at training a skeleton visual encoder such as UNIK~\cite{unik} in a self-supervised manner, without the need for action labels (see Fig.~\ref{fig:intro} (middle)). It is designed for the resulting visual encoder to be able to maximize the similarity of different skeleton sequences, that are obtained via data augmentation from the same original sequence, across large-scale datasets.
Unlike current methods~\cite{ding2022motion, selfrgb_2021_huang, transformer2022selfsupervised, selfrgb_2021_huang, Sun_2021_ICCV, ConNTU, Mao_2022_CMD, yang2022via} that perform contrastive learning for the video-level representations, we perform contrastive learning additionally on the frame space to finely maximize the per-frame similarities between the positive samples. Subsequently, the so-trained frame-level skeleton visual encoder is transferred and retrained on action segmentation datasets~\cite{Dai_2022_PAMI, Sigurdsson2016HollywoodIH}.
%\end{inparaenum}

To assess the performance of LAC, we train the skeleton visual encoder on the large-scale dataset Posetics~\cite{unik} and we evaluate the quality of the learned skeleton representations (see Fig.~\ref{fig:intro} (right)) by fine-tuning onto unseen action segmentation datasets (\eg, TSU~\cite{Dai_2022_PAMI}, Charades~\cite{Sigurdsson2016HollywoodIH}, PKU-MMD~\cite{liu2017pku}). Experimental analyses confirm that action composition and contrastive learning can significantly increase the expressive power of the visual encoder. The fine-tuning results outperform state-of-the-art accuracy.

In summary, the contributions of this paper include the following.
\begin{inparaenum}[(i)]
\item We introduce LAC, a novel generative and contrastive framework, streamlined to synthesize complex motions and improve the skeleton action representation capability.
\item In the generative step, we introduce a novel 
%and interpretive 
Linear Action Decomposition (LAD) mechanism to represent high-level motion features thanks to an orthogonal basis. The motions for multiple skeleton sequences can thus be linearly combined by latent space manipulation. 
\item In the contrastive learning step, we propose to learn the skeleton representations in both, video and frame space to improve generalization onto frame-wise action segmentation tasks.
\item We 
conduct experimental analysis and 
show that pre-training LAC on Posetics and transferring it onto an unseen target untrimmed video dataset represents a generic and effective methodology for action segmentation.

\end{inparaenum}
}
\section{Related Work}
\begin{figure*}[t]
\begin{center}

\includegraphics[width=1\linewidth]{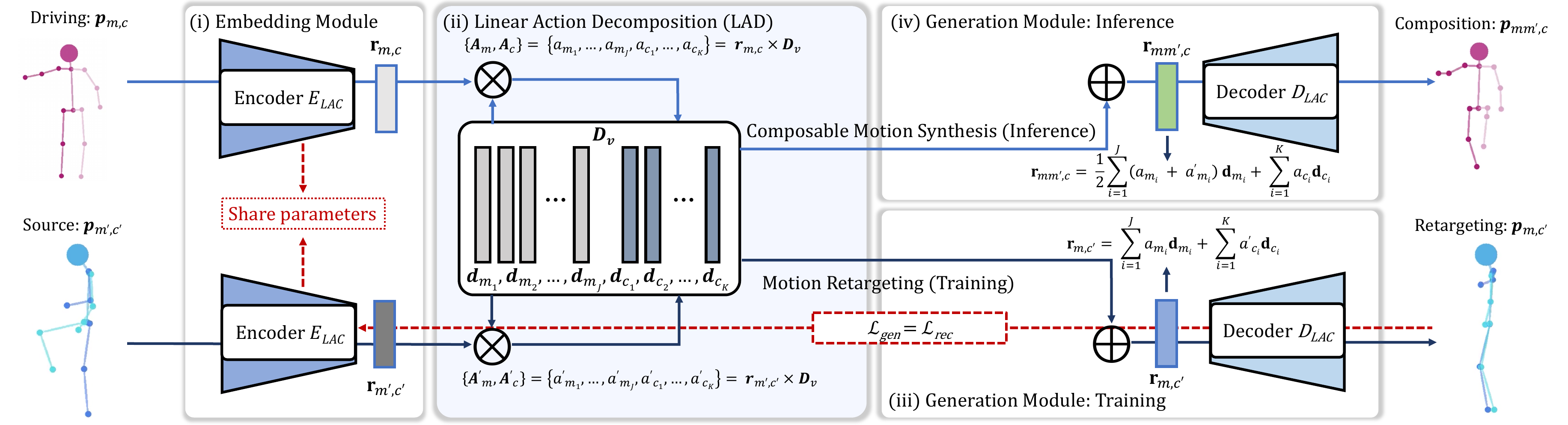}
\end{center}
   \vspace{-0.1cm}
   \caption{\textbf{Overview of the Composable Action Generation model in LAC.} The model consists of a visual encoder $\EM$ and a decoder $\D$. In the latent space, we apply Linear Action Decomposition (LAD) by learning a visual action dictionary $\mathbf{D}_v$, which is an orthogonal basis where each vector represents a basic `Motion'/`Static' transformation. Given a pair of skeleton sequences $\mathbf{p}_{m,c}$ and $\mathbf{p}_{m',c'}$, (i) their latent codes $\mathbf{r}_{m, c}$ and $\mathbf{r}_{m',c'}$ are embedded by $\EM$. (ii) Their projections $A_m$, $A_c$ and $A_{m'}$, $A_{c'}$ along $\mathbf{D}_v$ can be computed. The linear combination of $A_m$/$A_{m'}$ with corresponding directions in $\mathbf{D}_v$ constitutes the `Motion' features and similarly 
   %the linear combination of $A_c$ (or $A_{c'}$) with corresponding directions in $\mathbf{D}_v$ is considered as 
   the `Static' features can also be obtained. (iii) In the \textbf{training} stage, we leverage motion retargeting for learning the whole framework by swapping their `Motion' features and generating transferred motions. 
   %Specifically, the `Motion' features of the driving sequence $\mathbf{r}_{m}$ can be disentangled and regrouped with the `Static' features of $\mathbf{p}_{m',c'}$ to become the target code $\mathbf{r}_{m,c'}$. Finally, the target skeleton sequence $\mathbf{p}_{m,c'}$ is generated from $\mathbf{r}_{m,c'}$ by $\D$. 
   (iv) In the \textbf{inference} stage, 
   %as $D_v$ is learned to represent the visual 
   we adopt linear combination of $\mathbf{r}_{m}$ and $\mathbf{r}_{m'}$ to obtain the composable motion features $\mathbf{r}_{mm'}$ and the composable skeleton sequences can be generated.}
   %by $\mathbf{D}_v$ taking $\mathbf{r}_{mm'}$ and $\mathbf{r}_{c}$.} 
\vspace{-0.1cm} %0.33
\label{fig:overview}
\end{figure*}

\paragraph{Temporal Action Segmentation} focuses on per-frame activity classification in untrimmed videos. The main challenge has to do with how to model long-term relationships among various activities at different time steps. Current methods mostly focus on directly using untrimmed RGB videos. Since untrimmed videos usually contain thousands of frames, training a single deep neural network directly on such videos is quite expensive. Hence, to solve this problem efficiently, previous works proposed to use a two-step method. In the first step, a pre-trained feature extractor (\eg, I3D~\cite{Carreira_2017_CVPR}) is applied on short sequences to extract corresponding visual features. In the second step, action segmentation is modeled as a sequence-to-sequence (seq2seq) task to translate extracted visual features into per-frame action labels. Temporal Convolution Networks (TCNs)~\cite{lea2017temporal, dai2021pdan, zhangtqn} and Transformers~\cite{dai2022mstct} are generally applied in the second step due to their ability to capture long-term dependencies. 

Recently, few methods~\cite{Dai_2021_ICCV, Dai_2022_PAMI} started to explore using skeletons in this task, in order to benefit from multi-modality information. In such methods, a pre-trained Graph Convolutional Network (GCN) such as AGCN~\cite{2sagcn2019cvpr} is used as a visual encoder to obtain skeleton features in the first step. However, unlike in pre-trained I3D which has strong generalizability across domains, pre-trained AGCN is not able to provide high-quality features due to its laboratory-based pre-trained dataset NTU-RGB+D~\cite{Shahroudy2016NTURA}. We found that the performance significantly decreases when the pre-trained model 
is applied to more challenging real-world untrimmed skeleton videos datasets such as TSU~\cite{Dai_2022_PAMI} and Charades~\cite{Sigurdsson2016HollywoodIH}. The main issue is that the pre-trained visual encoder does not have a sufficient expressive power to extract the complex action features especially for composable actions that often occur in real-world videos. 

LAC differs from previous two-step methods. We propose a motion generative module to synthesize complex composable actions and to leverage such synthetic data to train a more general skeleton visual encoder~\cite{unik} which is sensitive to composable action. 
Unlike previous approaches, the pre-trained visual encoder in LAC 
%is pre-trained on synthetic composable skeletons which 
has stronger representation capability for skeleton sequences compared to previous two-step methods~\cite{Dai_2021_ICCV, Dai_2022_PAMI} using pre-trained AGCN. In such strategy, the model can be end-to-end refined on the action segmentation tasks without need for the second stage.

\vspace{-0.36cm}

\paragraph{Motion Retargeting} aims to transfer motion from sequence of target subject onto source subject, where the main challenge lies in developing effective mechanisms to disentangle motion and appearance. As one of the most important applications of video generation~\cite{Tulyakov:2018:MoCoGAN, Wang_2020_CVPR, WANG_2020_WACV, yu2022digan, stylegan_v, wang2021inmodegan}, previous image-based motion retargeting approaches explore to leverage structure representations such as 2D human keypoints~\cite{NKN_2018_CVPR, 2dmr, dance, yang2022via} and 3D human meshes~\cite{liu2019liquid,wang2021dance} as motion guidance. Recently, self-supervised methods~\cite{siarohin2019first, siarohin2021motion, wang2022latent} showed remarkable results on human bodies and faces by only relying on data without extract information. 

Skeleton-based methods~\cite{10.1145/3386569.3392462, Villegas_2021_ICCV, 2dmr, 10.1145/3386569.3392469} focus on transferring motion across skeletons of different shapes. Previous method~\cite{2dmr} showed that transferring motion across characters enforces the disentanglement of static and dynamic information in a skeleton sequence. While they have achieved good performance, such method is unable to compose different actions for creating novel actions. {Our method is different, we seek to learn an orthogonal basis in the feature space to represent the action distribution in a linear and discrete manner. 
In such a novel strategy, both static and dynamic features can be learned from a single encoder and skeleton sequences with complex motions are able to be synthesized by simply modifying the magnitudes along the basis. }

\vspace{-0.36cm}

\paragraph{Self-supervised Skeleton Action Representation} learning involves extracting spatio-temporal features from numerous unlabeled data.
Current methods~\cite{orvpe, ConNTU, guo2022aimclr, Mao_2022_CMD, yang2022via} adopt contrastive learning~\cite{tian2020cmc, non-para, He_2020_CVPR} as the pretext task to learn skeleton representations invariant to data augmentation. 
However, recent techniques~\cite{motionconsistency, colorization, orvpe, ConNTU, guo2022aimclr, Mao_2022_CMD, yang2022via} merge the temporal features by average pooling and conduct contrastive learning on top of the global temporal features for the skeleton sequences. Thus they may lose important information of complex actions particularly in the case of co-occurring actions~\cite{Dai_2022_PAMI, Sigurdsson2016HollywoodIH}. In our work, we extend the visual encoder and the contrastive module to finely extract per-frame features. We use contrastive loss for both sequence and frame, to make sure that the skeleton sequences are discriminative in both spaces. 
The skeleton visual encoder can have a strong representation ability for the sequence and also for each frame to better generalize to frame-wise action segmentation tasks.

\section{Proposed Approach}
LAC is composed of two modules (see Fig.~\ref{fig:intro}), a skeleton sequence generation module to synthesize the co-occurring actions and a self-supervised contrastive module to learn skeleton visual representations using the synthetic data. Subsequently, the skeleton visual encoder trained by the contrastive module can be transferred to downstream fine-grained action segmentation tasks. In this section, we introduce the full architecture and training strategy of LAC.

\subsection{Composable Action Generation}

In this work, we denote the static information of a skeleton sequence (\ie, `viewpoint', `subject body size', etc.) as `Static', while the temporal information (\ie, the dynamics of the `action' performed by the subject) as `Motion'. As shown in Fig.~\ref{fig:overview}, the generative module is an autoencoder, consisting of an encoder and a decoder for skeleton sequences. To disentangle `Motion' features from `Static' in a linear latent space, we introduce a Linear Action Decomposition mechanism to learn an action dictionary where each direction represents a basic high-level action for the skeleton encoding. We apply motion retargeting for training the autoencoder %
(\ie, transferring the motion of a driving skeleton sequence to the source skeleton sequence maintaining the source skeletons invariant in viewpoint and body size).
In the inference stage, the extracted `Motion' features from multiple skeleton sequences can be combined linearly and composable skeletons can be generated by the decoder. The input skeletons can be in 3D or 2D.

\vspace{-0.25cm}

\paragraph{Skeleton Sequence Autoencoder:} \label{sec:embedding}
The input skeleton sequence with `Static' $c$ and `Motion' $m$ is modeled by a spatio-temporal matrix, noted as $\mathbf{p}_{m,c}\in \mathbb{R}^{T \times V \times C_{in}}$. $T$, $V$, and $C_{in}$ respectively represent the length of the video, the number of body joints in each frame, and the input channels ($C_{in}=2$ for 2D data, or $C_{in}=3$ if we use 3D skeletons). 
As shown in Fig.~\ref{fig:overview} (i), LAC adopts an encoder $\EM$ to embed a pair of input skeleton sequences $\mathbf{p}_{m,c}$/$\mathbf{p}_{m',c'}$  into $\mathbf{r}_{m,c}$/$\mathbf{r}_{m',c'} \in \mathbb{R}^{T' \times C_{out}}$. $T'$ is the size of temporal dimension after convolutions and $C_{out}$ is the output channel size. 
To generate skeleton sequences, a skeleton sequence decoder $\D$ (see Fig.~\ref{fig:overview} a.(iii)) is used to generate new skeleton sequences from the representation space. The autoencoder is designed by multiple 1D temporal convolutions and upsampling to respectively encode and decode the skeleton sequence. We provide in Tab.~\ref{tab_arc} and Supplementary Material (Appendix) building details of $\EM$ and $\D$.

\vspace{-0.25cm}

\paragraph{Linear Action Decomposition:}
The goal of Linear Action Decomposition (LAD) is to obtain the `Motion' features on top of the encoded latent code of a skeleton sequence (see Fig.~\ref{fig:overview} a.(ii)). Our insight is that the high-level action of a skeleton sequence can be considered as a combination of multiple basic and independent `Motion' and `Static' transformations (\eg, raising hand, bending over) with their amplitude from a fixed reference pose (\ie, standing in the front view, see Fig.~\ref{fig:motion}). Hence, we explicitly model the basic `Static' and `Motion' transformations using a unified action dictionary for the encoded latent skeleton features. 
Specifically, we first pre-define a learnable orthogonal basis, noted as $\mathbf{D}_v = \{ \mathbf{d_m}_1, \mathbf{d_m}_2, ..., \mathbf{d_m}_J,  \mathbf{d_c}_1, \mathbf{d_c}_2, ..., \mathbf{d_c}_K \}$ with $J \in [1, C_{out})$ and $K=C_{out}-J$, where each vector indicates a basic `Motion'/`Static' transformation from the reference pose. Due to $\mathbf{D}_v$ entailing an orthogonal basis, any two directions $\mathbf{d_i}, \mathbf{d_j}$ follow the constraint:
\begin{equation}\label{Dictionary}
\small
<\mathbf{d_i}, \mathbf{d_j}>=\left\{\begin{matrix}
 0&i \neq j\\ 
 1&i = j.
\end{matrix}\right.
\end{equation}
We implement $\mathbf{D}_v  \in \mathbb{R}^{C_{out} \times C_{out}}$ as a learnable matrix and we apply the Gram-Schmidt algorithm during each forward pass in order to satisfy the orthogonality. Then, we consider the `Motion' features of $\mathbf{p}_{m,c}$, denoted as $\mathbf{r}_m$, as a linear combination between motion orthogonal directions in $\mathbf{D}_v$, and associated magnitudes (amplitude) $A_m=\{ a_{m1}, a_{m2}, ..., a_{mJ} \}$. Similarly, the `Static' features $\mathbf{r}_c$ are the linear combination between `Static' orthogonal directions in $\mathbf{D}_v$, and associated magnitudes $A_c=\{ a_{c1}, a_{c2}, ..., a_{cK} \}$. For $\mathbf{p}_{m',c'}$, we can obtain its decomposed components $\mathbf{r}_{m'}$, $\mathbf{r}_{c'}$ in the same way:
\begin{equation}\label{decompose}
\small
\begin{aligned}
    &\mathbf{r}_{m} = \sum_{i=1}^{J}a_{mi} \mathbf{d_{m}}_i, &\mathbf{r}_{c}= \sum_{i=1}^{K}a_{ci} \mathbf{d_{c}}_i,\\
    &\mathbf{r}_{m'} = \sum_{i=1}^{J}a'_{mi} \mathbf{d_{m}}_i,
    &\mathbf{r}_{c'} = \sum_{i=1}^{K}a'_{ci} \mathbf{d_{c}}_i.
\end{aligned}
\end{equation}
For the skeleton encoding $\mathbf{r}_{m, c}$/$\mathbf{r}_{m', c'}$, the set of magnitudes $A_m$/$A_m'$ and $A_c$/$A_c'$ can be computed as the projections of $\mathbf{r}_{m, c}$/$\mathbf{r}_{m', c'}$ onto $\mathbf{D}_v$, as Eq.~\ref{A}: 
\begin{equation}\label{A}
\small
\begin{aligned}
    &a_{mi} = \frac{<\mathbf{r}_{m, c} \cdot \mathbf{d_{m}}_i>}{\left\| \mathbf{d_m}_i\right\| ^2},&
    &a_{ci} = \frac{<\mathbf{r}_{m, c} \cdot \mathbf{d_c}_i>}{\left\| \mathbf{d_c}_i\right\| ^2},\\
    &a'_{mi} = \frac{<\mathbf{r}_{m', c'} \cdot \mathbf{d_m}_i>}{\left\| \mathbf{d_m}_i\right\| ^2},&
    &a'_{ci} = \frac{<\mathbf{r}_{m', c'} \cdot \mathbf{d_c}_i>}{\left\| \mathbf{d_c}_i\right\| ^2}.
\end{aligned}
\end{equation}
As $\mathbf{r}_{m, c}$ has the temporal dimension of size $T'$, for each `Motion' feature in the temporal dimension, we can obtain $T'\times$ sets of motion magnitudes $A_m$ to represent the temporal dynamics of $\mathbf{r}_{m}$. For $\mathbf{r}_{c}$, as static information, we firstly merge the temporal dimension of $\mathbf{r}_{m, c}$ by average pooling and then conduct the projection process to obtain a unified $A_c$. With such trained LAD, the decoder $\D$ can generate different skeleton sequences by taking an arbitrary combination of magnitudes $A_m$ and $A_c$ along their corresponding directions as input. The high-level action can thus be controlled by the manipulations in the latent space.
\begin{table}[t]
\centering
\begin{center}
\scalebox{0.85}{

\setlength{\tabcolsep}{0.7mm}{
\setlength{\arraycolsep}{0.5mm}{
\begin{tabular}{ c| c |c|c }
\hline
Stages  &$\EM$ &$\D$ &$\EV$\\

\hline
\hline
\multirow{2}*{Input} & 2D sequence& Rep.& 2D sequence   \\
 & [$T, 2V$]& [$T', ~160$]& [$T \times V, ~2$]\\

\hline

\multirow{2}*{1} 

&
\multirow{2}*{Conv$\left( \begin{array}{cc}
      8,& 64 
\end{array} \right)$}
&
Upsample(2)
& \multirow{2}*{
Conv$\left(\begin{array}{cc}
      1\times 1,&64 \\
      9\times 1,&64
\end{array} \right) \times 4$}
\\
& &
Conv$\left( \begin{array}{cc}
      7,& 128
\end{array} \right)$& \\
\hline

\multirow{2}*{2} 
&
\multirow{2}*{Conv$\left( \begin{array}{cc}
      8,& 96 
\end{array} \right)$}

& Upsample(2)&

\multirow{2}*{Conv$\left( \begin{array}{cc}
    1 \times 1,& 128 \\
    9 \times 1,& 128
     \end{array} 
     \right) \times 3$ }
\\
&& Conv$\left( \begin{array}{cc}
      7,& 64
\end{array} \right)$ &\\  

\hline
\multirow{2}*{3}
&
\multirow{2}*{Conv$\left( \begin{array}{cc}
      8,& 160 
\end{array} \right)$}

& Upsample(2)
&
\multirow{2}*{Conv$\left( \begin{array}{cc}
    1 \times 1,& 256 \\
    9 \times 1,& 256
     \end{array} 
     \right) \times 3$} 

\\
& &
Conv$\left( \begin{array}{cc}
      7,& 2V
\end{array} \right)$& \\
\hline
\multirow{1}*{4}
&  -&- &\multirow{1}*{S-GAP $(2 \times V,~256)$ }\\

%\hline
%\multirow{1}*{5}& -& -  &[$T', ~256$]\\
\hline
\multirow{2}*{Rep.} & \multirow{2}*{-} &\multirow{2}*{-} & $\EVF$: $[T,~256]$\\
 & & & $\EVS$: T-GAP to $[1,~256]$ \\
\hline
5& -&-& FC, Softmax\\
\hline
\multirow{2}*{Output}& \multirow{2}*{[$T', ~160$]} &2D sequence &\multirow{2}*{Per-frame Action Class}\\
& &[$T, ~2V$] &\\
\hline
\end{tabular}}}}
%}}
\end{center}
\vspace{-0.2cm}
\caption{\textbf{Main building blocks} of the autoencoder $\EM$, $\D$ and the skeleton visual encoder $\EV$ in LAC. We take the 2D sequence as example. The dimensions of kernels are denoted by $t \times s, c$ (2D kernels) and $t, c$ (1D kernels) for temporal, spatial, channel sizes. S/T-GAP, FC denotes temporal/spatial global average pooling, and fully-connected layer respectively. Rep. indicates the learned representation.}
\vspace{-0.3cm}
\label{tab_arc}
\end{table}

\vspace{-0.2cm}

\paragraph{Training (Motion Retargeting):}
We apply a general motion retargeting~\cite{2dmr} to train the generative autoencoder and ensure that `Motion' directions in LAD orthogonal basis $\mathbf{D}_v$ are `Static'-disentangled (see Fig.~\ref{fig:overview} (iii)). The main training loss function is the \textit{reconstruction loss}: $\mathcal{L}_{gen}$=$\mathcal{L}_{rec}$. 
%See Appendix for the details about optional triplet loss and velocity loss.
Reconstruction loss aims at guiding the network towards a high generation quality.
%at the global sequence level. 
The new retargeted (motion swapped) skeleton sequence with `Motion' $m$, and `Static' $c'$, noted as $\mathbf{p}_{m,c'}$ is generated from the recombined features, $\mathbf{r}_m+ \mathbf{r}_{c'}$. Similarly, $\mathbf{p}_{m',c}$ can also be generated by swapping the pair of sequences. The skeleton sequence generation can be formulated as $\mathbf{p}_{m,c'} = \D (\mathbf{r}_m+ \mathbf{r}_{c'})$ and $\mathbf{p}_{m',c} = \D (\mathbf{r}_{m'}+ \mathbf{r}_c)$. 
The reconstruction loss consists of two components: $\mathcal{L}_{rec} = \mathcal{L}_{self} + \mathcal{L}_{target}$.
Specifically, at every training iteration, the decoder network $\D$ is firstly used to reconstruct each of the original input samples $\mathbf{p}_{m,c}$ using its representation $\mathbf{r}_{m}+ \mathbf{r}_{c}$. This component of the loss is denoted as $\mathcal{L}_{self}$ and formulated as a standard autoencoder reconstruction loss (see Eq.~\ref{recloss1}).
\begin{equation}\label{recloss1}
\small
\begin{aligned}
\mathcal{L}_{self} = \mathbb{E} [\left\| \D (\mathbf{r}_m + \mathbf{r}_c)- \mathbf{p}_{m,c}\right\|^2],\\
\mathcal{L}_{target} = \mathbb{E} [\left\| \D (\mathbf{r}_m+ \mathbf{r}_{c'})- \mathbf{p}_{m,c'}\right\|^2].
\end{aligned}
\end{equation}
Moreover, at each iteration, the decoder is also encouraged to re-compose new combinations. 
As the generative module is trained on a synthetic dataset~\cite{mixamo} including the cross-character motion retargeting ground-truth skeleton sequences, we can explicitly apply the cross reconstruction loss $\mathcal{L}_{target}$ (see Eq.~\ref{recloss1}) through the generation. The same reconstruction losses are also computed for $\mathbf{p}_{m',c'}$.

\vspace{-0.35cm}

\paragraph{Inference (Motion Composition):}
As the trained LAD represents high-level motions in a linear space by the action dictionary, we can generate at the inference stage (see Fig.~\ref{fig:overview} (iv)) composable motions by the linear addition of `Motion' features encoded from multiple skeleton sequences. 
%To ensure that the generated motions are realistic, 
We use the average latent `Motion' features for the decoder to generate composable motions. 
We note that even if in some cases the combined motions may not be realistic, it can still help to increase the expressive power of the representation, which is important to express subtle details.
Taking the motion combination of the two sequences $\mathbf{p}_{m,c}$ and $\mathbf{p}_{m',c'}$ as an example, the skeleton sequences $\mathbf{p}_{mm',c}$ and $\mathbf{p}_{mm',c'}$ with the combined motions $m$ and $m'$ are generated as follows:  
\begin{equation}\label{inference}
\small
\begin{aligned}
    & \mathbf{p}_{mm',c} = \D \big( \frac{1}{2}(\mathbf{r}_m + \mathbf{r}_{m'}) + \mathbf{r}_c \big), \\
    & \mathbf{p}_{mm',c'} = \D \big( \frac{1}{2}(\mathbf{r}_m + \mathbf{r}_{m'}) + \mathbf{r}_{c'} \big).
\end{aligned}
\end{equation}
As skeleton sequences $\mathbf{p}_{mm',c}$ and $\mathbf{p}_{mm',c'}$ have the same composed motion but different `Static' (\eg, viewpoints), they can form a positive pair for self-supervised contrastive learning to train a transferable skeleton visual encoder for fine-grained action segmentation tasks in Sec.~\ref{sec:contrastive}.

\subsection{Self-supervised Skeleton Contrastive Learning}~\label{sec:contrastive}
In this section, we provide details of the self-supervised contrastive module of LAC.
We re-denote the generated composable skeleton sequence $\mathbf{p}_{mm',c}$ (in Sec.~\ref{sec:embedding}) as a query clip $q$ and multiple positive keys (\eg, the sequence $\mathbf{p}_{mm',c'}$), denoted as $k^{+}_1, ..., k^{+}_P$, can be generated by only modifying its `Static' magnitudes $A_{c}$ in the latent space. 
We follow the general contrastive learning method~\cite{He_2020_CVPR} based on the momentum encoder, to maximize the mutual information of positive pairs (\ie, the generated composable skeleton sequences with the same motion but different Statics), while pushing negative pairs (\ie, other skeleton sequences with different Motions) apart. %selected from the queue (memory)~\cite{He_2020_CVPR}) apart. 
Deviating from \cite{He_2020_CVPR}, the queue (memory)~\cite{He_2020_CVPR} stores the features of each frame for skeleton sequences and we propose to additionally enhance the per-frame representation similarity of positive pairs. The visual encoder can extract skeleton features that are globally invariant and also finely invariant to data augmentation and can generalize better to frame-wise action segmentation tasks.

\vspace{-0.3cm}
\paragraph{Skeleton Visual Encoder:}
To have a strong capability to extract skeleton spatio-temporal features, we adopt the recent topology-free skeleton backbone network UNIK~\cite{unik} as the skeleton visual encoder $\EV$ (see Tab.~\ref{tab_arc} and Appendix for details). 
To obtain the global sequence space, we adopt temporal average pooling layer to merge the temporal dimension of the visual representations, denoted as $\EVS(q), \EVS(k^{+}_1), ..., \EVS(k^{+}_P) \in \mathbb{R}^{C_{out} \times 1}$ (see Tab.~\ref{tab_arc}). 
Per-frame features can be obtained by $\EV$ before the temporal average pooling layer (see Tab.~\ref{tab_arc}) and denoted as $\EVF(q, \tau), \EVF(k^{+}_1, \tau), ..., \EVF(k^{+}_P, \tau) \in \mathbb{R}^{C_{out} \times T}$. 

\vspace{-0.3cm}
\paragraph{Contrastive Loss:}
We apply general contrastive InfoNCE loss~\cite{infonce} to train our visual encoder $\EV$ to encourage similarities between both sequence-level and frame-level representations of positive pairs, and discourage similarities between negative representations, denoted as $\EVS(k^{-}_1), ..., \EVS(k^{-}_N)$ in sequence space and $\EVF(k^{-}_1, \tau), ..., \EVF(k^{-}_N, \tau)$ in frame space. 
The InfoNCE~\cite{infonce} objective is defined as: $\mathcal{L}_{q} = \mathcal{L}_{q-s} + \mathcal{L}_{q-f}$, where
\begin{equation}\label{contras-loss-f}
\small
    \mathcal{L}_{q-s}  = - \mathbb{E} \Big( \log  \frac{\sum_{p=1}^P e^{\Sim\big ( \EVS (q), \EVS (k^{+}_p) \big ) }}{ \sum_{n=1}^N e^ {\Sim\big ( \EVS (q), \EVS(k^{-}_n) \big )  }} \Big),
\end{equation}
\begin{equation}\label{contras-loss-v}
\small
    \mathcal{L}_{q-f}  = - \mathbb{E} \Big( \log  \frac{\sum_{p=1}^P e^{\sum_{\tau=1}^T \Sim\big ( \EVF (q, \tau), \EVF (k^{+}_p, \tau) \big ) } }{ \sum_{n=1}^N e^ {\sum_{\tau=1}^T \Sim\big ( \EVF (q, \tau), \EVF(k^{-}_n, \tau) \big ) }} \Big),
\end{equation}
where $\tau$ represents the frame index in the temporal dimension of frame-level representations, $P$ represents the number of positive keys, $N$ denotes the number of negative keys (we use $P=4$ and $N=65,536$ for experiments), and the similarity is computed as:
\begin{equation}\label{sim}
\small
    \Sim (x, y) =  \frac{\phi (x) \cdot \phi (y)}{\left\| \phi (x) \right\| \cdot \left\| \phi (y) \right\|} \cdot \frac{1}{Temp},
\end{equation}
where $Temp$ refers to the temperature hyper-parameter~\cite{non-para}, and $\phi$ is a learnable mapping function (\eg, a MLP projection head~\cite{study2021}) that can substantially improve the learned representations.

\vspace{-.4cm}
\paragraph{Transfer-Learning for Action Segmentation:}
For transferring the visual encoder on downstream tasks, we attach $\EVF$ to a fully-connected layer followed by a Softmax Layer to predict per-frame actions. The output size of each fully-connected layer depends on the number of action classes (see Tab.~\ref{tab_arc}). Then, we re-train the visual encoder $\EV$ with action labels. For processing long sequences, we adopt a sliding window to extract features for a temporal segment and use Binary Cross Entropy loss to optimize the visual encoder step by step. In this way, $\EV$ can be re-trained end-to-end instead of pre-extracting features for all frames. In the inference stage, we combine the predictions of all the temporal sliding windows in an online manner~\cite{jcrnn}.
\section{Experiments and Analysis}

In this section, we conduct extensive experiments to evaluate LAC on both generation and action segmentation tasks. Firstly, we study the generalization ability of LAC by quantifying the performance improvement obtained by transfer-learning on target action segmentation datasets (\ie, \text{Toyota Smarthome Untrimmed}, \text{Charades} and \text{PKU-MMD}) after pre-training on the large-scale dataset \text{Posetics}. 
Secondly, we evaluate the quality of the skeleton sequences generated by LAC using the synthetic dataset \text{Mixamo}.
Finally, we provide an exhaustive ablation study. 
See Appendix 
for implementation details, limitation discussion and additional studies, \eg, computational cost analysis. 

\begin{table}[t]
\centering

\begin{center}
\scalebox{0.9}{
\setlength{\tabcolsep}{1.mm}{
\begin{tabular}{ l c c c c }
\hline
\multirow{2}*{\textbf{Methods}}& \multirow{2}*{\textbf{Mod.}}& \multicolumn{2}{c}{\textbf{TSU}}&\multirow{1}*{\textbf{Charades}} \\
&&{\text{ CS(\%)}} & {\text{CV(\%)}} & {\text{mAP(\%)}}\\

\hline
\hline
\rowcolor{mygray}\text{TGM~\cite{TGM}}&RGB &\text{26.7}&- & 13.4\\

\rowcolor{mygray}\text{PDAN~\cite{dai2021pdan}}&RGB &\text{32.7} & -& 23.7\\
%\rowcolor{mygray}\text{Coarse-Fine~\cite{kahatapitiya2021coarse}}&RGB &\text{-}&- &25.1 \\
\rowcolor{mygray}\text{SD-TCN~\cite{Dai_2022_PAMI}}&RGB &\text{29.2}&18.3 &21.6 \\
\rowcolor{mygray}\text{MS-TCT~\cite{dai2022mstct}}&RGB &\text{33.7} &- &25.4 \\

\hline

\text{Bi-LSTM~\cite{graves2005framewise}}&Skeleton &\text{17.0} &14.8 &8.2 \\
\text{TGM~\cite{TGM}}&Skeleton &\text{26.7} &13.4 &9.0 \\
\text{SD-TCN~\cite{Dai_2022_PAMI}}&Skeleton &\text{26.2}&22.4 &9.8 \\

\hline
\textbf{LAC-unsup (Ours)}&Skeleton & \textbf{34.1}&\textbf{22.8} &\textbf{22.3}\\
\text{LAC-sup (Ours)}&Skeleton & \textbf{36.8}&\textbf{23.1} &\textbf{25.6}\\
\hline
\end{tabular}}}
\end{center}
\vspace{-0.2cm}
\caption{Frame-level mAP on TSU and Charades for comparison with SoTA action segmentation methods. RGB-based results (top) are shown for reference. Mod.: Modality.}
\vspace{0.1cm}
\label{tab_sota1}
\end{table}

\begin{table}[t]
\centering

\begin{center}
\scalebox{0.9}{
\setlength{\tabcolsep}{.8mm}{
\begin{tabular}{ l c c c c }
\hline
\multirow{2}*{\textbf{Methods}}& \multirow{2}*{\textbf{Mod.}}& \multicolumn{3}{c}{\textbf{PKU-MMD} mAP@IoU} \\
&&{\text{ 0.1(\%)}} & {\text{0.3(\%)}} & {\text{0.5(\%)}}\\

\hline
\hline
%\rowcolor{mygray}Deep RGB & RGB& 50.7& 32.3& 14.7\\
%\rowcolor{mygray}Qin and Shelton & RGB& 65.0& 51.0& 29.4\\
\rowcolor{mygray}GRU-GD~\cite{GRU-GD} & RGB &82.4& 81.3& 74.3\\
\rowcolor{mygray}SSTCN-GD~\cite{Dai_2021_ICCV} & RGB&83.7 &82.1 &76.5\\
\rowcolor{mygray}Augmented-RGB~\cite{Dai_2021_ICCV} & RGB&86.3 &84.5 &81.1\\

\hline
JCRRNN~\cite{jcrnn} &Skeleton& 45.2& \text{-} &32.5 \\
Convolution Skeleton~\cite{liu2017pku} & Skeleton&49.3& 31.8 &12.1 \\
Skeleton boxes~\cite{skeletonbox}& Skeleton& 61.3& \text{-} & 54.8\\
%Wang and Wang& Skeleton& 84.2& \text{-}&\text{-}\\
Hi-TRS~\cite{hirachical2022}&Skeleton &-& \text{-} &67.3 \\
Window proposal~\cite{windowproposal} &Skeleton &92.2& \text{-} &90.4 \\

\hline
\textbf{LAC-unsup (Ours)}&Skeleton & \textbf{91.8}&\textbf{90.2} &\textbf{88.5}\\
\text{LAC-sup (Ours)}&Skeleton & \textbf{92.6}&\textbf{91.4} &\textbf{90.6}\\
\hline
\end{tabular}}}
\end{center}
\vspace{-0.2cm}
\caption{Event-level mAP on PKU-MMD CS at IoU thresholds of 0.1, 0.3 and 0.5 for comparison with SoTA methods. RGB-based results (top) are shown for reference. Mod.: Modality.}
\vspace{-0.3cm}
\label{tab_sota2}
\end{table}

\begin{table*}[t]
\centering

\begin{center}
%\resizebox{225pt}{63pt}{ 
\scalebox{0.91}{
%0.5
\setlength{\tabcolsep}{1.5mm}{
\begin{tabular}{  l c c c c c c c }

\hline
\multirow{2}*{\textbf{Methods}}&\multirow{2}*{\textbf{Pre-training}} & \multirow{2}*{\textbf{Training data}}
  & \multicolumn{2}{c}{\textbf{Toyota Smarthome Untrimmed}} & \multicolumn{2}{c}{\textbf{PKU-MMD} (IoU=0.1)}& {\textbf{Charades}}\\

& & & \text{CS(\%)} &\text{CV(\%)}& {\text{ CS(\%)}} & {\text{CV(\%)}} & {\text{mAP(\%)}} \\
\hline
\hline

\text{Random init.~\cite{unik}}&Scratch  & 5\% &\text{8.5}& \text{6.8}& 57.4 &59.5  &8.8 \\

%\text{Supervised}& 5\% &\textbf{}& \textbf{}& \textbf{} & & &  \textbf{} \\
\textbf{Self-supervised}& Posetics w/o labels &  5\% &\textbf{25.2}& \textbf{15.6} &\textbf{73.9} &\textbf{75.4} & \textbf{12.6} \\

\hline
\text{Random init.~\cite{unik}}& Scratch&  10\% &\text{12.9}& \text{9.5}&  66.4& 68.1& 9.3  \\
%\text{Supervised}& 10\% &\textbf{}& \textbf{}& \textbf{} & & &  \textbf{} \\
\textbf{Self-supervised}& Posetics w/o labels & 10\% &\textbf{29.0}& \textbf{17.9}& \textbf{79.8} &\textbf{81.1} &\textbf{17.4}  \\

\hline
\end{tabular}}}

\end{center}
\vspace{-0.2cm}
\caption{Transfer learning results by \textbf{fine-tuning} on all benchmarks of Toyota Smarthome Untrimmed, PKU-MMD and Charades with randomly selected \textbf{5\% (top)} and \textbf{10\% (bottom)} of labeled training data. }
\vspace{0.1cm}
\label{tab_fewer}
\end{table*}

\begin{table*}[t]
\centering

\begin{center}
%\resizebox{225pt}{63pt}{ 
\scalebox{0.91}{
%0.5
%\setlength{\tabcolsep}{1.1mm}{
\begin{tabular}{  l c c c c c c c c c c }

\hline
\multirow{2}*{\textbf{Methods}}
& \multirow{2}*{\textbf{Pre-training}}& \multicolumn{3}{c}{\textbf{Toyota Smarthome Untrimmed}} & \multicolumn{3}{c}{\textbf{PKU-MMD} (IoU=0.1)}& \multicolumn{2}{c}{\textbf{Charades}}\\
%\cline{2-9} 
& &\text{\#Params}& \text{CS(\%)} &\text{CV(\%)} &\text{\#Params}& {\text{ CS(\%)}} & {\text{CV(\%)}} &\text{\#Params}& {\text{mAP(\%)}} \\
\hline
\hline
%\multicolumn{1}{l}{\textbf{Occlusion-robustness Evaluation}}\\
\text{Random init.}&Scratch & 13.1K&\text{8.1}& \text{6.9}  &13.3K &11.8  & 12.4& 40.2K &6.1 \\
%\text{3D-ResNets (RGB)}&&\textbf{}& \textbf{}& \textbf{} & & & & \textbf{} \\
%\text{Pr-ViPE}& &\textbf{}& \textbf{}& \textbf{} & & & & \textbf{} \\
%\text{CV-MIM}& &\textbf{}& \textbf{}& \textbf{} & & & & \textbf{} \\
%\text{3s-CrosSCLR}&&\textbf{}& \textbf{}& \textbf{} & & & & \textbf{} \\
%\text{OR-VPE}&7.97K&\textbf{42.7}& \textbf{18.1}& \textbf{32.4} & & & 3.85K& \textbf{78.5} \\
\text{Supervised}& Posetics w/ labels&  13.1K&\text{20.8}& 18.3 &  \text{13.3K} &61.8 &62.4 &  40.2K& \text{14.3} \\
\textbf{Self-supervised}& Posetics w/o labels&  13.1K&\textbf{18.5} & \textbf{16.6} & 13.3K &\textbf{55.2} &\textbf{58.8} & 40.2K&\textbf{12.7} \\

\hline
%\text{Previous SoTA}&-&- &\text{63.6~\cite{Ryoo2020AssembleNetAM}}& \text{$\dagger$43.8}~\cite{das2020vpn}&\text{54.6~\cite{unseenview}} & - &\text{38.0~\cite{shift_2020_CVPR}}& 67.0~\cite{shift_2020_CVPR} &- &\text{$\dagger$98.7~\cite{multitask}} \\
Random init. & Scratch&3.45M &28.2 &11.0 & 3.45M &86.5  &92.9 &3.45M &18.6  \\
%\text{Pr-ViPE}& &\textbf{}& \textbf{}& \textbf{} & & & & \textbf{} \\
%\text{CV-MIM}& &\textbf{}& \textbf{}& \textbf{} & & & & \textbf{} \\
%\text{3s-CrosSCLR}&&\textbf{}& \textbf{}& \textbf{} & & & & \textbf{} \\

\text{Supervised}& Posetics w/ labels &3.45M&\text{36.8}& \text{23.1} & 3.45M& \text{92.6} &94.6 &3.45M &\text{25.6}\\
%\text{OR-VPE}&3.45M&\textbf{56.3}& \textbf{24.6}&\textbf{59.0} & 3.45M& \textbf{93.3}& 3.45M& \textbf{84.5} \\
\textbf{Self-supervised}& Posetics w/o labels &3.45M&\textbf{34.1}& \textbf{{22.8}} & 3.45M &\textbf{91.8} &\textbf{93.9} &3.45M &\textbf{22.3} \\

\hline
%\text{Previous SoTA}&-&- &\text{}& \text{} &\text{} & - &\text{}&  &- &\text{} \\
%\textbf{LAC (Ours)}&Posetics w/o labels &-&\textbf{}& \textbf{}&\textbf{} & - &\textbf{}&\textbf{} &- &\textbf{} \\

%\hline
\end{tabular}}

\end{center}
\vspace{-0.2cm}
\caption{Transfer-learning results by \textbf{linear evaluation (top)} and \textbf{fine-tuning (bottom)} on Toyota Smarthome Untrimmed, PKU-MMD and Charades with self-supervised pre-training on Posetics. Results with supervised pre-training are also reported for reference.}
\vspace{-0.25cm}
\label{tab_tr}
\end{table*}  
%and \textbf{previous state-of-the-art (bottom)}

\subsection{Datasets and Evaluation Protocols}
\noindent\textbf{Posetics}~\cite{unik} contains 142,000 real-world trimmed video clips from Kinetics-400~\cite{Carreira_2017_CVPR} with corresponding 2D and 3D skeletons. We use Posetics to pre-train the contrastive model of LAC with skeleton data and we study the transfer-learning on skeleton-based action segmentation. 
\vspace{0.15cm}

\noindent\textbf{Toyota Smarthome Untrimmed (TSU)}~\cite{Dai_2022_PAMI} is a large-scale real-world dataset for daily living action segmentation. It contains densely annotated long-term composite activities where up to 5 actions can happen at the same time in a given frame. We only use the provided 2D skeleton data~\cite{Yang_2021_WACV} for the experiments. For evaluation, we report \textit{per-frame} mAP (mean Average Precision) as~\cite{dai2021pdan, dai2022mstct} following the cross-subject (CS) and cross-view (CV) evaluation protocols.

\vspace{0.15cm}

\noindent\textbf{Charades}~\cite{Sigurdsson2016HollywoodIH} is a real-world dataset containing fine-grained activities similar to TSU. It provides only raw video clips without skeleton data. In this work, 
we use the 2D skeleton data (2D coordinates) estimated by the toolbox~\cite{Yang_2021_WACV}.
We report \textit{per-frame} mAP on the localization setting of the dataset. For sake of reproducibility, we will release the estimated skeleton data on Charades.

\vspace{0.15cm}

\noindent\textbf{PKU-MMD}~\cite{liu2017pku} is a basic untrimmed video dataset recorded in the laboratory setting. We use only the official 3D skeleton data. As this dataset is not densely labeled, we report the \textit{event-based} mAP for fair comparisons by applying a post-processing~\cite{GRU-GD} on the frame-level predictions to get the action boundaries.

\vspace{0.15cm}
\noindent\textbf{Mixamo}~\cite{mixamo} is a 3D animation collection, which contains elementary actions and various dancing moves. 
We use such a synthetic dataset
for training and evaluating the generation module in LAC prior to contrastive learning on Posetics. 

\subsection{Evaluation on Temporal Action Segmentation}
In this section, we evaluate the transfer ability of LAC by both \textit{linear evaluation} (\ie, by training only the fully-connected layer while keeping frozen the backbone) and \textit{fine-tuning evaluation} (\ie, by refining the whole network) on three action segmentation datasets \text{TSU, PKU-MMD} and \text{Charades} with self-supervised pre-training on \text{Posetics}. We also report the results with supervised pre-training for reference (\ie, we use the generated composable skeletons and the combined action labels for pre-training).

\vspace{-0.4cm}
\paragraph{Linear Evaluation:}
Tab.~\ref{tab_tr} (top) shows the linear results on the three datasets. This evaluates the effectiveness of transfer-learning with fewer parameters (only the classifier is trained) compared to training directly on the target datasets from scratch (random initialization). The results suggest that the weights of the model can be well pre-trained without action labels, providing a strong transfer ability
(\eg, +10.4\% on TSU CS and +6.6\% on Charades) 
and the pre-trained visual encoder is generic enough to extract meaningful action features from skeleton sequences.

\vspace{-0.4cm}
\paragraph{Fine-tuning:}
Tab.~\ref{tab_tr} (bottom) shows the fine-tuning results, where the whole network is re-trained. 
The self-supervised pre-trained model also performs competitively compared to supervised pre-trained models. From these results we conclude that collecting a large-scale trimmed skeleton dataset, without the need of action annotation, can be beneficial to downstream fine-grained tasks for untrimmed videos (\eg, +5.9\% on TSU CS and +11.8\% on CV). 

\vspace{-0.4cm}

\paragraph{Training with fewer labels:}
In many real-world applications, labeled data may be lacking, which makes it challenging to train models with good performance. To evaluate LAC in such cases, we transfer the visual encoder pre-trained on Posetics onto all the tested datasets by fine-tuning with only 5\% and 10\% of the labeled data. As shown in Tab.~\ref{tab_fewer}, without pre-training, the accuracy of the visual encoder~\cite{unik} significantly decreases.
In contrast, LAC with prior action representation learning achieves good performance on all three datasets in such setting. 

\vspace{-0.4cm}

\paragraph{Comparison with SoTA:}
We compare our fine-tuning results to other SoTA skeleton-based approaches~\cite{graves2005framewise, TGM, Dai_2022_PAMI, jcrnn, skeletonbox, liu2017pku, hirachical2022, windowproposal} on the real-world datasets TSU and Charades (see Tab.~\ref{tab_sota1}) and also laboratory dataset PKU-MMD (see Tab.~\ref{tab_sota2}). As previous approaches are based on supervised pre-training on large-scale datasets~\cite{NTU-120, Carreira_2017_CVPR}, we also report our supervised results. The results in Tab.~\ref{tab_sota1} show that LAC, even with self-supervised pre-training, outperforms all previous skeleton-based approaches~\cite{graves2005framewise, TGM, Dai_2022_PAMI} with supervised pre-training on our main target real-world datasets in a large margin (\eg, +7.4\% on TSU CS and +12.5\% on Charades). It suggests that composable motions are important to increase the expressive power of the visual representation and the end-to-end fine-tuning can benefit downstream tasks. Even if PKU-MMD does not contain composable actions, the performance is still slightly improved by learning a fine-grained skeleton representation. The results using RGB data are also reported for reference. 
The TSU and Charades datasets contain many object-oriented actions that are difficult to identify using skeleton data only. However, even in the absence of the object information, LAC surprisingly achieves better accuracy compared to all SoTA RGB-based methods~\cite{dai2021pdan, Dai_2022_PAMI, dai2022mstct, TGM, Dai_2021_ICCV}. We deduce that training the visual encoder end-to-end is more effective compared to using two-step processing. Moreover, skeletons can always be combined with RGB data by multi-modal fusion networks~\cite{Dai_2021_ICCV, das2021vpn+} to further improve the performance.

\begin{figure}
\begin{center}
\includegraphics[width=1\linewidth]{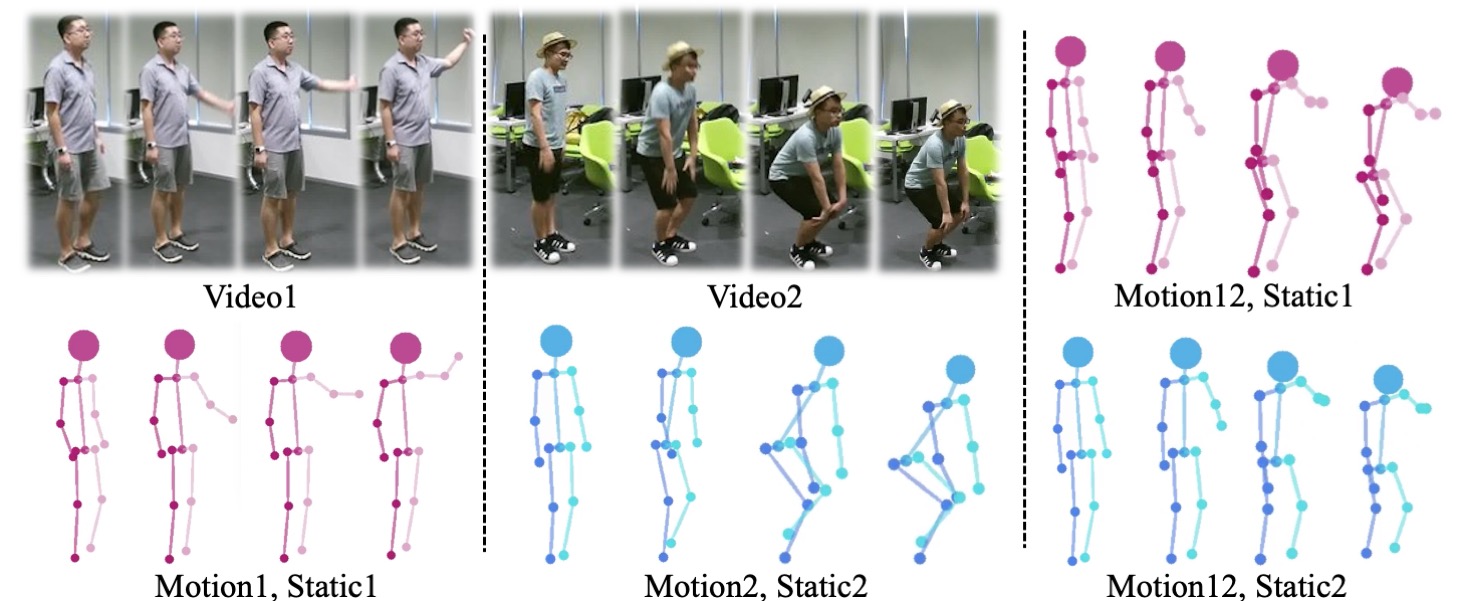}
\end{center}
   \vspace{-0.2cm}
   \caption{ \textbf{Motion composition visualization.} The input pair of videos and corresponding skeleton sequences (left) have simple motions. The generated skeleton sequences (right) are composed by both motions while keeping their respective viewpoint and body size (`Static') invariant.}
\vspace{0.1cm}
\label{fig:composition}
\end{figure}

\begin{figure}
\begin{center}

\includegraphics[width=1\linewidth]{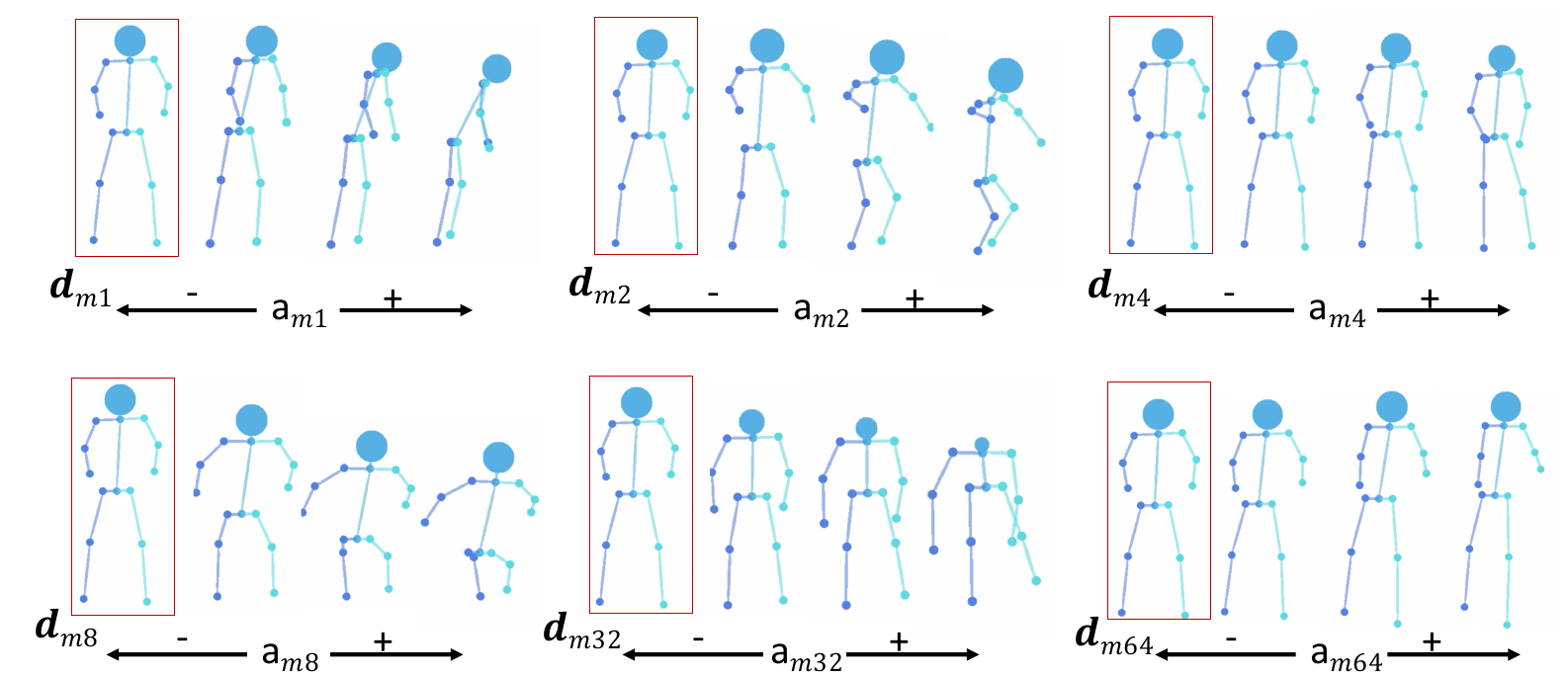}
\end{center}
   \vspace{-0.2cm}
   \caption{ \textbf{Linear manipulation of six `Motion' directions} in $\mathbf{D}_v$ on a skeleton sequence. Results indicate that each direction represents a meaningful motion transformation from a `reference pose' marked in red (\eg, $\mathbf{d_m}_8$ for squat, $\mathbf{d_m}_{32}$ for bending over).}
\vspace{-0.cm}
\label{fig:motion}
\end{figure}

\subsection{Evaluation on Action Generation.}
As the generative model with LAD represents our main novelty for addressing the action segmentation challenges, we evaluate here the generation quality of LAC.
\vspace{-0.4cm}

\paragraph{Quantitative Comparison:} The generation model of LAC is trained on the \text{Mixamo} dataset to have an action composition ability before the contrastive learning. We compare the motion retargeting accuracy on this dataset. Specifically, we randomly split training and test sets on this dataset and follow the same setting and protocol described in~\cite{2dmr, yang2022via}. We firstly explore how many directions (\ie, the values of $J$ and $K$) are required in the proposed action dictionary $\mathbf{D}_v$. We empirically test four different values for $J$ from 16 to 144. From results reported in Tab.~\ref{tab_mr}, we observe that when using 128 directions (out of all $dim$=160 directions) for `Motion', the model achieves the best reconstruction accuracy and outperforms SoTA methods~\cite{NKN_2018_CVPR, 2dmr, yang2022via}. Hence, we set $J$=128 and $K$=32 for all other experiments.

\vspace{-0.4cm}

\paragraph{Motion Direction Interpretation and Visualization:} We visualize an example of motion composition inference of two videos. Fig.~\ref{fig:composition} demonstrates that `Static' and `Motion' are well disentangled and the high-level motions can be effectively composed by decoding the linear combination of both latent `Motion' components learned by the proposed LAD.
To further understand what each direction in $\mathbf{D}_v$ represents, we proceed to visualize $\mathbf{d_{m}}_i$.
We generate skeletons for a single input skeleton sequence using its disentangled `Static' features $\mathbf{r}_{c}$ combined by different $\mathbf{r}_{m}$ respectively obtained by a linearly grown ${a_{m}}_i$ on its corresponding `Motion' directions $\mathbf{d_{m}}_i$ (see Fig.~\ref{fig:motion} for visualization of six directions), where other magnitudes on directions except $\mathbf{d_{m}}_i$ are set to 0. We find that each direction represents a basic high-level motion transformation (\eg, $\mathbf{d_{m}}_{32}$ represents bending over) and the corresponding magnitude represents the range of the motion. All motion transformations start from a fixed `reference pose', regardless of original motions of the input skeleton sequences. Such a `reference pose' can be considered as a normalized form of the given skeleton sequence.
In such a learning strategy, complex motions can be combined and the motion diversity can be controlled in an interpretive way by latent space manipulation. More real-world examples with different viewpoints are provided in the Appendix.

\begin{table}[t]
\centering

\begin{center}
\scalebox{.92}{
\begin{tabular}{ l c  }
\hline
\textbf{Methods}& \textbf{Mean Square Error}\\

\hline
\hline
\text{NKN~\cite{NKN_2018_CVPR}} &\text{1.51} \\
\text{MotionRetargeting2D~\cite{2dmr}} &\text{0.96}  \\
\text{ViA~\cite{yang2022via}}  & \text{0.86}\\
\hline

%\text{LAC w/o $\mathbf{D}_v$ (Ours)}  & \text{}\\
\text{LAC w/ $\mathbf{D}_v$ (Ours)}  \\
\text{~~~~size ${J}= 16$, ${K}= 144$}  & \text{1.23}\\
\text{~~~~size ${J}= 32$, ${K}= 128$}  & \text{1.02} \\
\text{~~~~size ${J}= 64$, ${K}= 96$}  & \text{0.88} \\
~~~~size \textbf{${J}= 128$, ${K}= 32$}  & \textbf{0.82}\\
\text{~~~~size ${J}= 144$, ${K}= 16$}  & \text{0.85} \\
\hline
\end{tabular}}
\end{center}
\vspace{-0.2cm}
\caption{Quantitative comparisons of LAC to other SoTA motion retargeting methods on the Mixamo dataset. }
\vspace{-.1cm}
\label{tab_mr}
\end{table}

\subsection{Ablation Study}

To understand the contribution of the two individual components of LAC, we conduct ablation experiments on our main target fine-grained dataset TSU, with self-supervised pre-training and fine-tuning protocol. 
\vspace{-0.4cm}

\begin{table}[t]
  \centering
     
     \scalebox{0.92}{
        \begin{tabular}{  l c c  c }
        
        \hline
        \multirow{1}*{\textbf{Toyota Smarthome Untrimmed}}
        &\multirow{1}*{\text{CS (\%)}} & 
           \multicolumn{1}{c}{\text{CV (\%)}}\\
           \hline
           \hline
        \text{L0: Base: w/o LAC} & 29.8& \text{13.8}\\
        \text{L1: +Motion Composition} & & \text{}\\
        ~~\text{Number of motions=2} & \text{33.8}& 21.9 \\
        ~~\text{Number of motions=3} & \text{32.1}& 21.1\\
  %      ~~\text{$\rho$=4} & & \text{}\\
        \text{L2: +Frame-level Contrast} &  &\textbf{} \\
        ~~\text{Temporal sample rate=2} &34.0& \text{22.5}\\
        ~~\text{Temporal sample rate=4} & \textbf{34.1}& \textbf{22.8}\\
        ~~\text{Temporal sample rate=8} &33.7 & \text{22.0}\\
        \hline
        \end{tabular}}
        \vspace{0.15cm}
     \caption{mAP on Toyota Smarthome Untrimmed CS and CV for showing impacts of two types of hyper-parameter for modulating the generated skeleton sequences.}
     \label{tab_ablation}
     \vspace{-0.25cm}
\end{table}

\paragraph{Impact of Action Composition:}
We start from a baseline model~\cite{unik} that is pre-trained on the trimmed dataset (\ie, Posetics) in a general contrastive learning strategy~\cite{He_2020_CVPR} without using composable motions and frame-level contrast for action segmentation. The results in Tab.~\ref{tab_ablation} (see L0) suggest that the visual encoder has a weak capability to learn features on top of an untrimmed skeleton sequence without learning a composable action representation.
We then perform the self-supervised training on Posetics (in only the video space) with composable motions from different number of motions. As daily living videos contain in average two co-occurring actions~\cite{Dai_2022_PAMI}, combining motions from two skeleton sequences in the pre-training stage can significantly improve the representation ability of the visual encoder and generalize better to real-world untrimmed action segmentation tasks (see Tab.~\ref{tab_ablation} L1). Such number can simply be changed to adapt to different target datasets.

\vspace{-0.4cm}
\paragraph{Impact of Frame-wise Contrast:}
To validate that frame-wise contrastive learning can further improve the fine-grained action segmentation tasks, we additionally maximize the per-frame similarity between the positive samples. We also select different uniform temporal sampling rates to reduce the redundant computational cost instead of using all the frames. The results in Tab.~\ref{tab_ablation} L2 suggest that frame-wise contrast with uniformly sampling every 4 frames is the most effective to improve the action segmentation accuracy.

\subsection{Further Discussion}

\paragraph{Transfer Learning vs. Self Pre-training:}
Our target is to train a generic skeleton encoder that can fit different downstream tasks. 
Hence, similar to current RGB-based methods using large-scale dataset such as Kinetics~\cite{Carreira_2017_CVPR, k700} for pre-training, our model is pre-trained 
on the large-scale Posetics dataset to learn a generic skeleton representation. Such representation can be transferred onto different downstream tasks without the need for individual pre-training. This is a very effective practice for action segmentation models.
To demonstrate the advantage of transfer-learning and to further compare LAC with SoTA methods, we here compare LAC with SoTA methods~\cite{TGM, Dai_2022_PAMI} in Tab.~\ref{tab_self} with self pre-training, \ie,
\textit{solely self-supervised pre-training the encoder on the tested dataset} (on TSU, PKU-MMD CS-IoU@0.1 and Charades) using the proposed contrastive module without additional data and without action labels.
The results show that, without extra training data, LAC can still outperform previous models~\cite{TGM, Dai_2022_PAMI}, as in the second stage, LAC adopts end-to-end fine-tuning to refine the visual encoder, which is more effective than using temporal modeling on the pre-extracted features~\cite{TGM, Dai_2022_PAMI}. Moreover, current untrimmed datasets are not large enough, the generated actions have less diversity, so the representation ability of the skeleton encoder is less impressive than pre-training on Posetics. 
\begin{table}[t]
\centering

\begin{center}
\scalebox{.92}{
       \setlength{\tabcolsep}{1.3mm}{
        \begin{tabular}{  l c c c  } 
        \hline
        \textbf{Dataset}& {\text{TGM}~\cite{TGM}} & {\text{SD-TCN}~\cite{Dai_2022_PAMI}}& {\textbf{LAC (Ours)}}
     \\
        \hline
        \hline
        \textbf{TSU-CS(\%)} & 25.6 & 24.4& \textbf{33.2} \\
        \textbf{TSU-CV(\%)} &13.9 & 20.8& \textbf{21.7}\\
        \textbf{Charades(\%)} & 9.1& 8.7&\textbf{21.4}\\
       \textbf{PKU-MMD(\%)}&87.3 &87.5 &\textbf{91.0}\\
        \hline
        \end{tabular}}}
\end{center}
\vspace{-0.2cm}
\caption{Fine-tuning results (\ie, Frame-level mAP on TSU and Charades and Event-level mAP on PKU-MMD) with individual pre-training only on the target action segmentation datasets for further comparison with SoTA methods.}
\vspace{-0.25cm}
\label{tab_self}
\end{table}  

\vspace{-0.cm}

\section{Conclusion}
In this work, we present LAC, a novel self-supervised action representation learning framework for the setting of skeleton action segmentation. 
We show that high-level motions of skeleton sequences can be learned and linearly combined using an orthogonal basis in the latent space.
%and composable motion can be generated by a linear combination of magnitudes along the orthogonal basis projected from multiple skeleton sequences. 
Moreover, we augment a contrastive learning module to better extract frame-level features, in addition to the generated composable skeleton sequences. Our experimental analysis confirms that a skeleton visual encoder that extracts such skeleton representation is able to boost downstream action segmentation tasks.
%especially with composable actions.
%on large-scale datasets (\eg, Kinetics-400) boosts the classification accuracy, when transferred onto downstream target datasets (\eg, Toyota Smarthome, UCF101 and HMDB51). This is specifically the case in fine-grained action classification tasks.
%
%
Future work will extend our generative approach to RGB videos, in order to improve the capturing of the object information, which can be crucial and complementary to the skeleton-based model.
\vspace{-0.2cm}

\paragraph{Acknowledgements:} This work was supported by Toyota Motor Europe (TME) and the French government, through the 3IA Cote d’Azur Investments In the Future project managed by the National Research Agency (ANR) with the reference number ANR-19-P3IA-0002. %This work was granted access to the HPC resources of IDRIS under the allocation AD011012798R1. 
This work was performed using HPC resources from GENCI–IDRIS (Grant AD011012798R1).

{\small
\balance
\bibliographystyle{ieee_fullname}
\bibliography{egbib}
}

\end{document}